\documentclass{article}
\usepackage{spconf,amsmath,graphicx}
\usepackage{multirow}
\usepackage{amsfonts}
\usepackage{hyperref}

\title{MULTI-MODALITY GENERATIVE ADVERSARIAL NETWORKS WITH TUMOR CONSISTENCY LOSS FOR BRAIN MR IMAGE SYNTHESIS}
\name{Bingyu Xin$^{1,2,*}$ \qquad Yifan Hu$^{3}$ \qquad Yefeng Zheng$^{3}$ \qquad Hongen Liao$^{1,2}$   \thanks{* xinby17@mails.tsinghua.edu.cn}}

\address{$^{1}$ Department of Biomedical Engineering, \\School of Medicine, Tsinghua University, Beijing 100084, China\\
$^{2}$ Institute of Biomedical Engineering, \\Graduate School at Shenzhen, Tsinghua University, Shenzhen 518055, China\\
$^{3}$ Tencent Youtu Lab, Shenzhen 518057, China}

\begin{document}
%
\maketitle

\begin{abstract}
Magnetic Resonance (MR) images of different modalities can provide complementary information for clinical diagnosis, but whole modalities are often costly to access. Most existing methods only focus on synthesizing missing images between two modalities, which limits their robustness and efficiency when multiple modalities are missing. To address this problem, we propose a multi-modality generative adversarial network (MGAN) to synthesize three high-quality MR modalities (FLAIR, T1 and T1ce) from one MR modality T2 simultaneously. The experimental results show that the quality of the synthesized images by our proposed methods is better than the one synthesized by the baseline model, pix2pix. Besides, for MR brain image synthesis, it is important to preserve the critical tumor information in the generated modalities, so we further introduce a multi-modality tumor consistency loss to MGAN, called TC-MGAN. We use the synthesized modalities by TC-MGAN to boost the tumor segmentation accuracy, and the results demonstrate its effectiveness.
\end{abstract}
\begin{keywords}
Image synthesis, Generative Adversarial Networks, Brain tumor segmentation, Multi-modality
\end{keywords}

\section{Introduction}
\label{sec:intro}

MR images are widely used in neurology and neurosurgery. MR images can provide exquisite details of brain, spinal cord and vascular anatomy, which can be used for segmentation of tumors and organs at risk. Different MR modalities, such as T1, T2, T1 with contrast enhanced (T1ce) and Fluid Attenuation Inversion Recover (FLAIR), emphasize different types of biological information and tissue properties. For instance, T2 and FLAIR MRI indicate differences in tissue water relaxational properties and T1ce MRI shows pathological take-up of contrast agents in tumor area \cite{menze2014multimodal}. Provided with complementary modalities, clinicians can know better about their patients, and automatic segmentation algorithms can also perform better to accelerate the process of improved clinical diagnosis and radiotherapy treatment planning. However, acquisition of different modalities is time consuming, costly and sometimes may be impossible for the lack of specific imaging equipment. Missing modalities means missing corresponding information, which may result in  misdiagnosis or severe degradation on segmentation performance. Therefore, different methods have been explored to synthesize the missing modalities.

     Recently, generative adversarial networks (GANs) \cite{goodfellow2014generative,mirza2014conditional} have aroused great research interest to synthesize missing modalities from existing modalities. In the field of medical image synthesis \cite{yi2019generative}, DCGAN \cite{radford2015unsupervised} can be used to generate new data examples. CycleGAN \cite{zhu2017unpaired} is often used to synthesize for unpaired images. When paired images are available, pix2pix \cite{isola2017image} is a preferred network since it uses the paired information in the dataset. Frid-Adar et al. \cite{frid2018synthetic} use DCGAN to synthesize new data as a data augmentation method for lesion classification. Dar et al. \cite{dar2019image} adopt pix2pix and CycleGAN to realize cross-modality translation between T1 and T2 brain MR images. Yu et al. \cite{yu2019ea} integrate edge information to improve the MR image synthesis results of pix2pix.
     
     MR images often have several modalities, however, both pix2pix and CycleGAN can only synthesize images from one modality to another. For example, if we have T2 MR images and want to synthesize other three modalities, three independent pix2pix or CycleGAN networks need to be trained, which would be inefficient and unstable. For unpaired multi-domain data, StarGAN \cite{choi2018stargan} introduces domain labels to CycleGAN and enable a single network to translate an input image to any desired target domain. Inspired by StarGAN, for paired multi-modality MR images, we propose to introduce modality labels to pix2pix so that a single modality T2 can be translated by a single network to any target modalities. We find our proposed method is similar to UAGAN \cite{yuan2019unified} and CollaGAN \cite{lee2019collagan}, however, UAGAN follows StarGAN to perform any-to-any image synthesis, and CollaGAN uses three existing modalities to synthesize one missing modality, which is on the contrary to ours. In addition, a recent research shows that the synthesized images by GANs are always biased by distribution matching loss \cite{cohen2018distribution}, we also find similar phenomenon in our experiments, the tumor shape of brain FLAIR images generated by pix2pix isn’t consist with original T2 images. To tackle this problem, Jiang et al. \cite{jiang2018tumor} introduce a tumor-aware loss to preserve the tumor information. In this work, we propose a multi-modality tumor consistency loss to our network to maintain the tumor mapping among multiple modalities.

       Our work has three contributions. First, we add modality labels into the generator and discriminator of the original pix2pix to enable a unified network to synthesize multiple modalities, and the experimental results demonstrate that the proposed method can generate higher quality images. Second, we introduce a multi-modality tumor consistency loss to preserve the tumor information during the process of MR image synthesis, which further improves the image structural similarity of the synthesized modalities to the source MR images. Third, we use our proposed TC-MGAN to generate FLAIR, T1 and T1ce brain MR images from T2 modality, then use the synthesized modalities to boost the tumor segmentation accuracy. The experiments show that the tumor segmentation results of our method are better than the one only using T2 modality. 
\begin{figure}[t]

\begin{minipage}[b]{1.0\linewidth}
  \centering
  \centerline{\includegraphics[width=7.0cm]{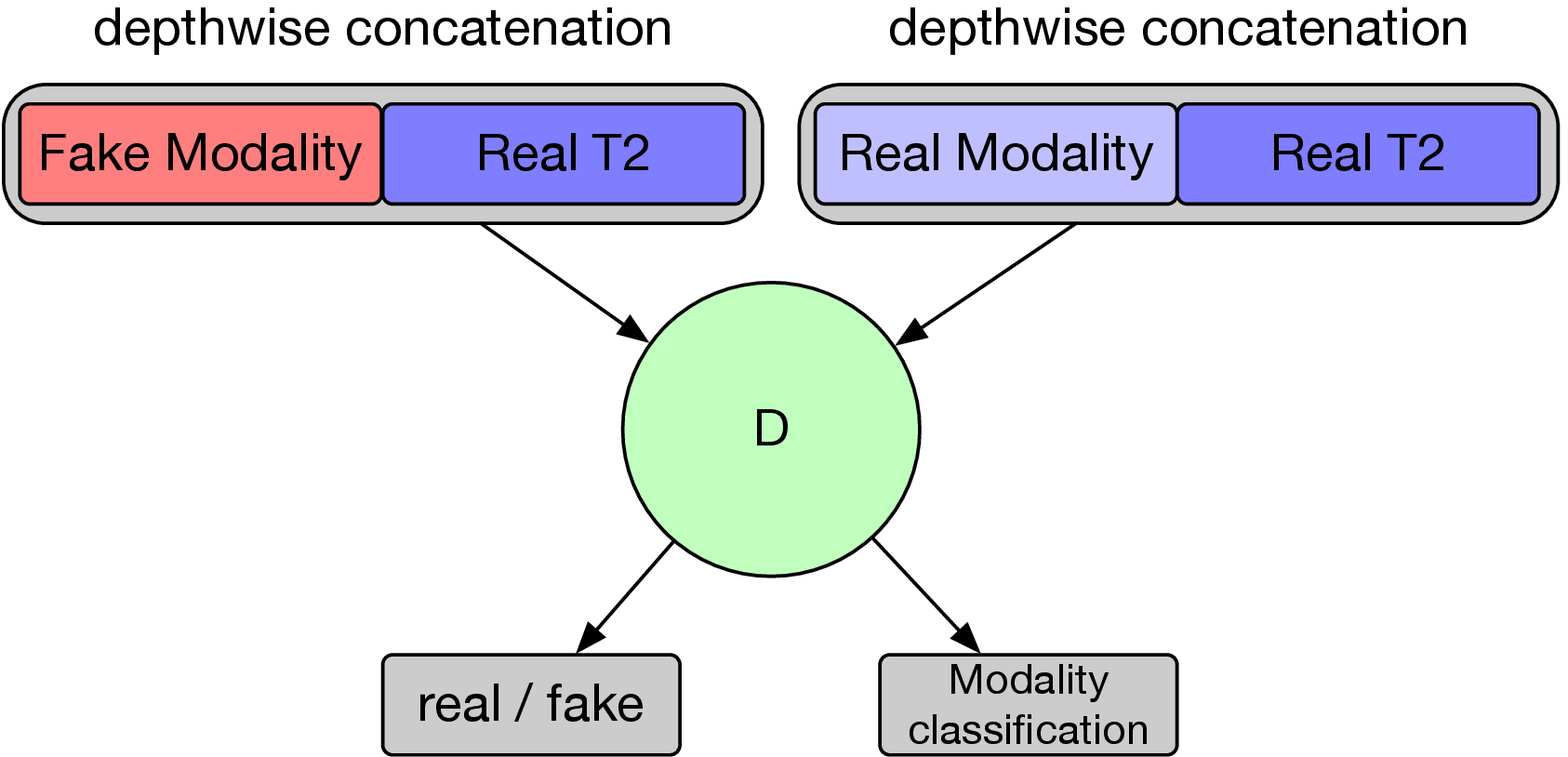}}
  \centerline{(a) Train Discriminator}\medskip
\end{minipage}
\begin{minipage}[b]{1.0\linewidth}
  \centering
  \centerline{\includegraphics[width=8.5cm]{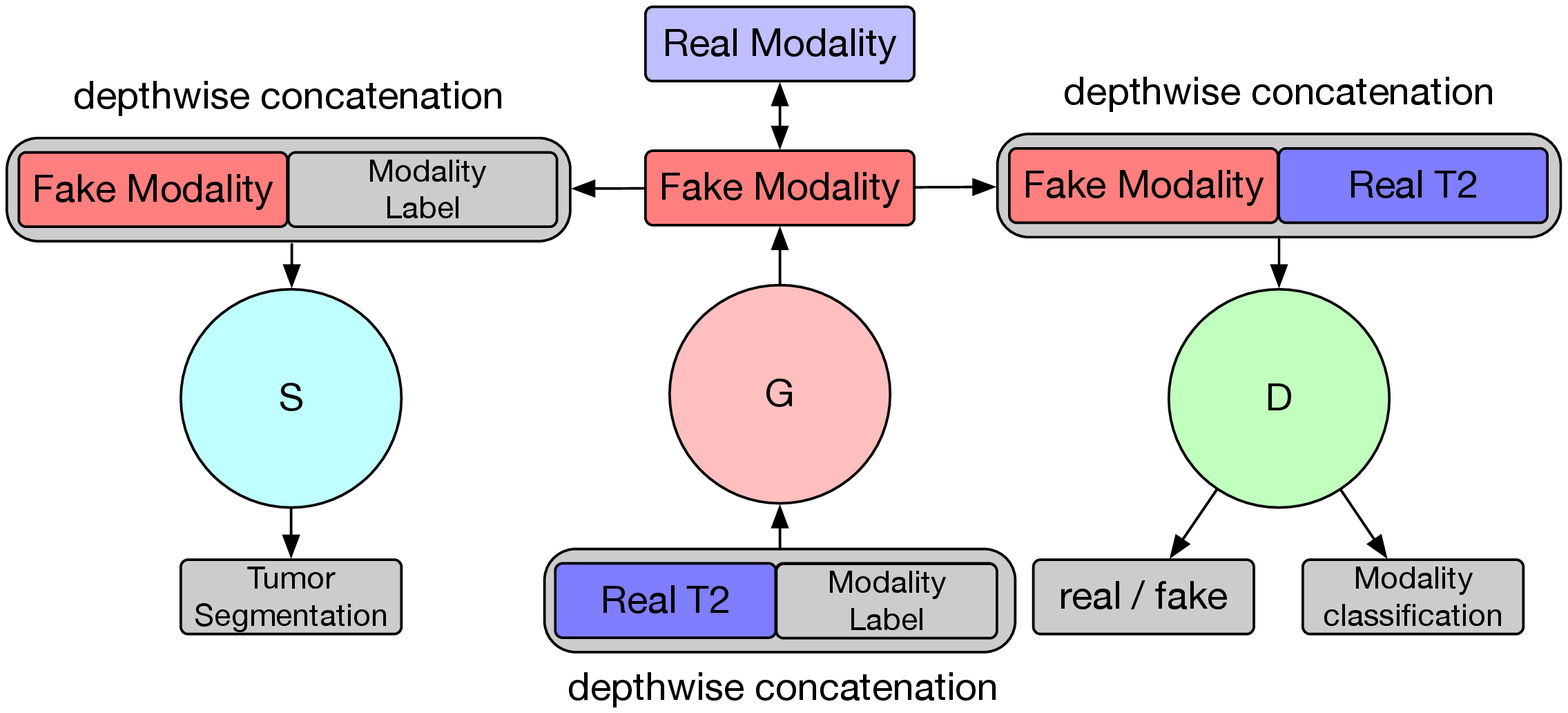}}
  \centerline{(b) Train Generator}\medskip
\end{minipage}
%
\caption{Overview of TC-MGAN. (a) discriminator $D$ learns to distinguish between real and fake images and classify modalities of the input images. (b) generator $G$ learns to generate realistic target modality to fake $D$ and segmentor $S$ forces  $G$ to focus on cross modality mapping in tumor area.}
\label{fig:model}
\end{figure}

\section{PROPOSED METHODS}
\label{sec:format}
\subsection{Pix2pix}
\label{ssec:subhead_prop}
Pix2pix is the baseline model in this paper. Pix2pix consists of two modules, the generator $G$ and the discriminator $D_{src}$. In the process of training, the generator $G$ tries to generate samples $G(x)$ resembling to real samples $y$ to trick the discriminator $D_{src}$, and the discriminator $D_{src}$ tries to distinguish between synthesized samples and real samples. The pix2pix is optimized by this zero-sum game, which ultimately enables the generator to generate realistic samples. The objective function of pix2pix is defined as:
\begin{align}
    \mathop{min}\limits_G \mathop{max}\limits_D \mathcal{L}(G,D_{src}) & =\mathbb{E}_{x,y}[\log{(D_{src}(x,y))}] \notag \\ & +\mathbb{E}_{x }[\log{(1-D_{src}(x,G(x)))}]\notag \\ & + \lambda_{l1}\mathcal{L}_{l1}(G(x),y) 
\end{align}

where the first two items are adversarial loss, $\mathcal{L}_{l1}$ is L1 loss to ensure the pixel-wise similarity between synthesized images and real images. $\lambda_{l1}$ is a hyper-parameter to balance its weight.
\subsection{Modality Labels}
\label{ssec:subhead_modal}
In order to control the generator $G$ to synthesize any target modalities, we introduce modality labels $c$ to the input of the generator, thus we can specify the modality of synthesized images by adjusting the values of $c$. We concatenate the modality label $c$ to the source image $x$ as the input of the generator, so the synthesized image is $G(x,c)$. To ensure the modality label $c$ can take effect in generator $G$, we need the discriminator $D$ not only to distinguish real images from fake images, but also classify the modality of the input image. To this end, the modality classification loss is defined as:
\begin{align}
    \mathcal{L}_{cls}^r= \mathbb{E}_{x,y,c}[-\log{D_{cls}(c|(x,y_c))}]
\end{align}
\begin{align}
    \mathcal{L}_{cls}^f= \mathbb{E}_{x,c}[-\log{D_{cls}(c|(x,G(x,c)))}]
\end{align}
where superscript $r$ denotes real image input and $f$ denotes fake image input.

\subsection{Multi-Modality Tumor-consistency Loss}
\label{ssec:subhead_multi}
For medical image synthesis, it is important to ensure that the tumor information in source modality can be well preserved in target modality, otherwise it may lead to severe misdiagnosis of patients based on hallucinated tumor information. Therefore, a multi-modality tumor segmentation network is introduced to guide the translation process. We concatenate the corresponding modality label $c$ to the synthesized image as the input of the segmentation network $S$. For ease of training, $S$ is pre-trained on the same training dataset and the parameters are fixed when training the GAN model. We calculate the Dice loss to compare the segmentation result with the tumor label map, which is then used to optimize the generator through back-propagation. The multi-modality tumor consistency loss is defined as:
\begin{align}
    \mathcal{L}_{seg}= \mathbb{E}_{x,c}[S(G(x,c),gt)]
\end{align}
where $gt$ is the ground truth of the tumor segmentation map. It is worth noting that different modality naturally reflects distinct characteristics of human anatomy. $S$ is not proposed to force the target modality to have the same tumor appearance as the source modality, but to constrain the generator to focus on learning the cross-modality mapping in tumor area. 

\begin{figure}[b]

\begin{minipage}[b]{1.0\linewidth}
  \centering
  \centerline{\includegraphics[width=8.5cm]{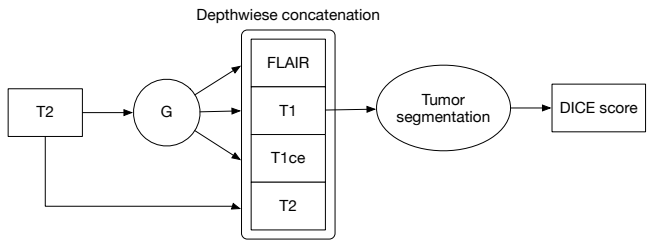}}
\end{minipage}

\caption{The pipeline to boost tumor segmentation results using synthesized modalities.}
\label{fig:pipeline}
\end{figure}

\subsection{Implement Details}
\label{ssec:subhead_imp}
To stabilize the training process and improve the generalization of the GAN model, we adopt objective function and gradient penalty proposed in WGAN \cite{gulrajani2017improved} to all the GAN models in our experiments, it is defined as:
\begin{align}
    \mathcal{L}_{gp}= 
    \mathbb{E}_{x,\hat{x}} [(\left\| \nabla{\hat{x}} D_{src}(x,\hat{x}) \right\|_2-1)^2]
\end{align}
Where $\hat{x}$ is sampled uniformly along a straight line between a pair of a real and a synthesized image. The final objective function of the discriminator in proposed TC-MGAN is defined as:
\begin{align}
    \mathcal{L}_{D}&= -\mathbb{E}_{x,y,c}[D_{src}(x,y_c)]+\mathbb{E}_{x,c}[D_{src}(x,G(x,c))] \notag \\
    &+\lambda_{cls}\mathcal{L}_{cls}^r+\lambda_{gp}\mathcal{L}_{gp}
\end{align}
The final objective function of the generator is defined as:
\begin{align}
\mathcal{L}_{G}&= -\mathbb{E}_{x,c}[D_{src}(x,G(x,c))] \notag \\
    &+\lambda_{cls}\mathcal{L}_{cls}^f+\lambda_{l1}\mathcal{L}_{l1}+\lambda_{seg}\mathcal{L}_{seg}
\end{align}
where hyper-parameters $\lambda_{cls}$, $\lambda_{gp}$, $ \lambda_{l1}$, $ \lambda_{seg}$ are used to balance different terms. The proposed TC-MGAN is shown in Figure \ref{fig:model}. During each training iteration, we randomly generate target modality label $c$ for the generator $G$ to synthesize corresponding target modality and for segmentation network $S$ to segment the tumor area for the synthesized modality. Code is available at \url{https://github.com/hellopipu/TC-MGAN}.

\section{EXPERIMENTAL RESULTS}
\label{sec:pagestyle}

\subsection{Dataset and Experimental Setting}
\label{ssec:subhead_data}
We use BRATS18 \cite{menze2014multimodal} multi-modal brain tumor dataset for our experiments. The dataset contains 285 subjects with four modalities of co-registered MR images: FLAIR, T1, T1ce and T2, with the image size $240\times 240\times 155$. We use T2 as the source modality to synthesize other three modalities since T2 is a widely used modality in clinical diagnosis. The dataset is divided into 100 subjects for training GAN network and multi-modality segmentation network S, 100 subjects for training segmentation networks for comparing the tumor segmentation accuracy improvement when using synthesized modality, and the rest 85 subjects for testing the quality of synthesized images and the tumor segmentation accuracy.

      \begin{figure}[t]

\begin{minipage}[b]{1.0\linewidth}
  \centering
  \centerline{\includegraphics[width=8.5cm]{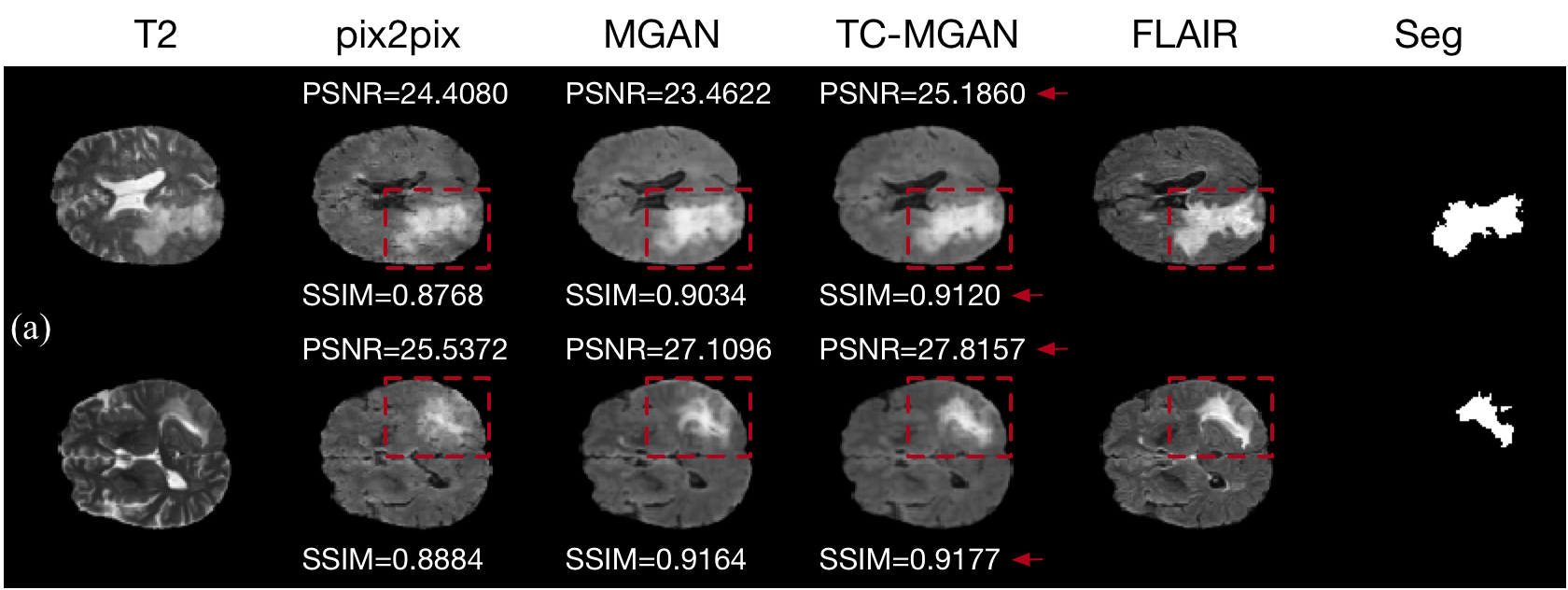}}
\end{minipage}
\begin{minipage}[b]{1.0\linewidth}
  \centering
  \centerline{\includegraphics[width=8.5cm]{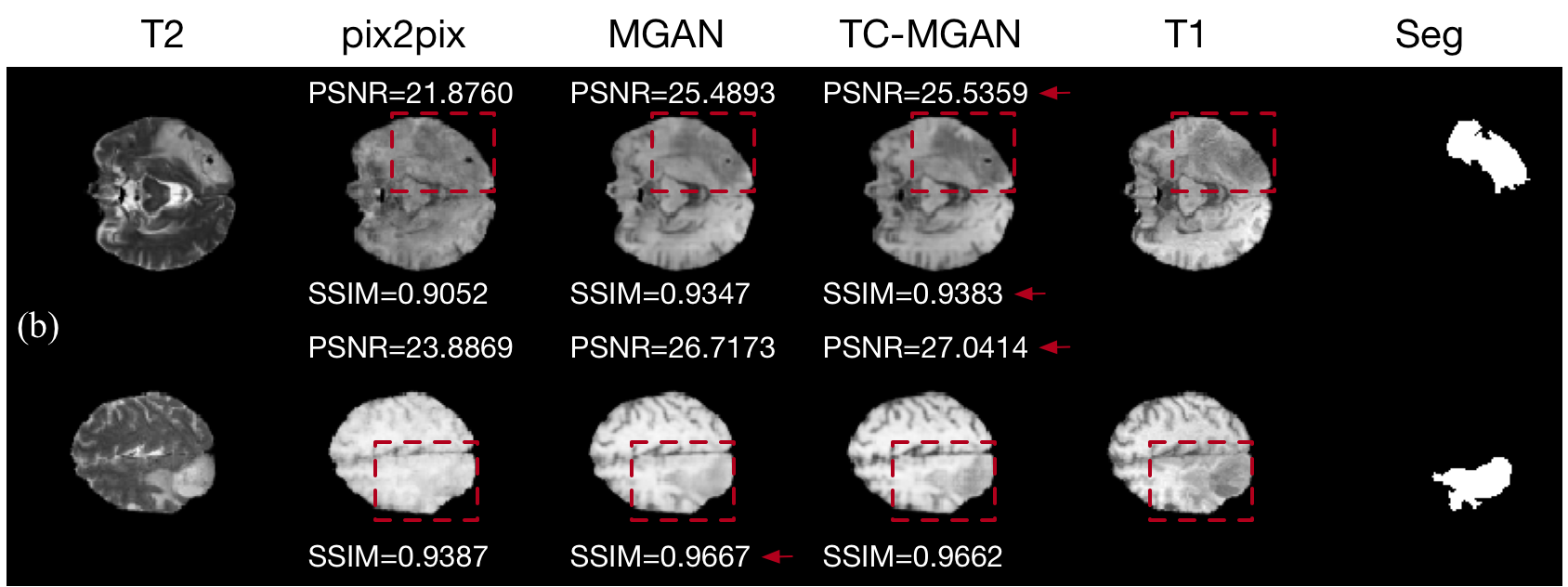}}
\end{minipage}
\hfill
\begin{minipage}[b]{1.0\linewidth}
  \centering
  \centerline{\includegraphics[width=8.5cm]{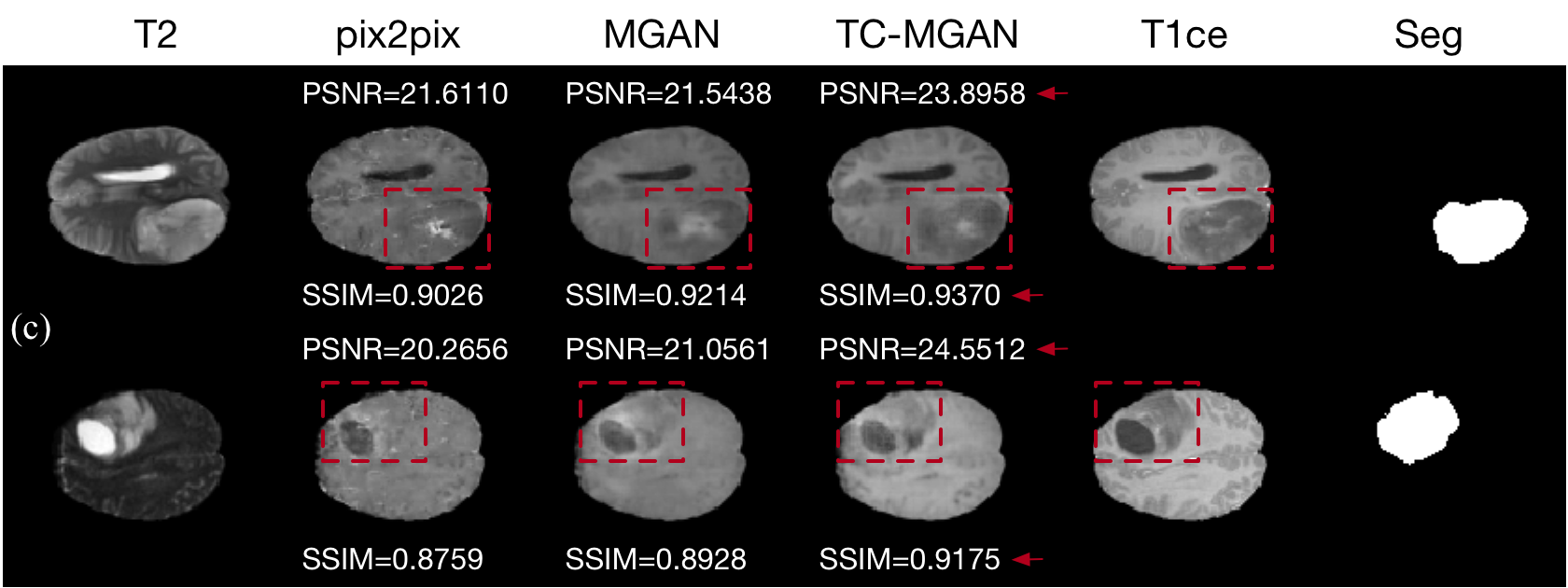}}
\end{minipage}
\caption{Example synthesized images from T2 to other three modalities. Our methods can well preserve the tumor information and have better quality than pix2pix. (a) T2 to FLAIR (b) T2 to T1 (c) T2 to T1ce.}
\label{fig:result}
\end{figure}

\begin{table*}[t]
\renewcommand\arraystretch{1.5}
\small
\centering
\caption{Quality comparison of synthesized images.}
\label{psnr}

\def\temptablewidth{1.0\textwidth}
{\rule{\temptablewidth}{1pt}}  

\begin{tabular*}{\temptablewidth}{@{\extracolsep{\fill}}c|lll|lll}

\multirow{2}{*}{\textbf{Methods}} & \multicolumn{3}{c|}{\textbf{PSNR}}                                             & \multicolumn{3}{c}{\textbf{SSIM}}        \\                                              \cline{2-7} 
                                  & \multicolumn{1}{c}{FLAIR} & \multicolumn{1}{c}{T1} & \multicolumn{1}{c|}{T1ce} & \multicolumn{1}{c}{FLAIR} & \multicolumn{1}{c}{T1} & \multicolumn{1}{c}{T1ce} \\ \hline
3 *pix2pix& 24.0345$\pm $3.1225 &   25.2793$\pm $3.6721  &   25.1622$\pm $3.1622   &  0.9110$\pm $0.0444  &0.9421$\pm $0.0355 & 0.9260$\pm $0.0359\\ \hline
MGAN& 24.6176$\pm $3.2926 &   \textbf{25.8468}$\pm $\textbf{3.8419}  &     \textbf{25.7269}$\pm $\textbf{3.8608} &  0.9256$\pm $0.0387 &0.9520$\pm $0.0327 & 0.9389$\pm $0.0319 \\ \hline
TC-MGAN&  \textbf{24.6416}$\pm $\textbf{3.4277} &  25.7784$\pm $3.9372  &   25.6822$\pm $3.9470    &  \textbf{0.9267}$\pm $\textbf{0.0386}  &0.9520$\pm $0.0326& 0.9390$\pm $0.0322 \\ \hline
TC-MGAN bw & 24.6164$\pm $3.4460	&25.8126$\pm $3.9763 &	25.6962$\pm $3.9776& 0.9266$\pm $0.0382&	\textbf{0.9528}$\pm $\textbf{0.0322}&	\textbf{0.9399}$\pm $\textbf{0.0319}\\ 
\end{tabular*}
{\rule{\temptablewidth}{1pt}}
\end{table*}

The networks are all 2D networks, we filter out all axial slices whose pixel number in brain area is less than 2000 and resize the rest axial slices from $240\times 240$ to $128\times 128$, and then linearly scale the original intensity to [-1, 1]. In the experiments, $\lambda_{l1}$, $\lambda_{cls}$, $\lambda_{seg}$,$ \lambda_{gp}$ are set to 0.1, 10, 50, and 100, respectively. Adam optimizer with a batch-size of 64 is applied to minimize the objectives. For all segmentation networks, we conduct 30 epochs to train the model, with learning rate set to 0.001. For all synthesis networks, we conduct 100 epochs to train, learning rate set to 0.0002 with the first 30 epochs, after 30, 60, 90 epochs, learning rate set to 0.0001, 0.00005, 0.00001, respectively. The baseline model in our experiments is pix2pix, we compare it with our proposed methods, MGAN and TC-MGAN, which all share the same architecture, hyper-parameters, learning rate and all use gradient penalty to stabilize training for sake of fair comparison. We use Unet \cite{ronneberger2015u} as the architecture for all segmentation networks. In addition, Zhang et al \cite{zhang2018translating} proposed to train GAN model combined with an auxiliary segmentation network, while our method trains the GAN model with the parameters of segmentation network fixed, in order to compare the two training strategies in our experiments, TC-MGAN with the loss of segmentation network backward (TC-MGAN bw) is also included in our experiments.

\subsection{Evaluation Measures}
\label{ssec:subhead_eva}

We employ two measurements to evaluate the synthesis performance of the proposed MGAN, TC-MGAN, TC-MGAN bw and pix2pix in comparison: peak signal-to-noise ratio (PSNR) and structural similarity index (SSIM). Besides, we also compare the DICE score improvements for tumor segmentation by using the synthesized images generated by our methods and by pix2pix. The pipeline for boosting the segmentation result is shown in Figure \ref{fig:pipeline}. We use the generator $G$ to synthesize three missing modalities from the source modality T2, then all four modalities are concatenated as the input of the tumor segmentation network.

\subsection{Image Synthesis Results}
\label{ssec:subhead_syn}
Figure \ref{fig:result} shows the synthesized samples from T2 by different methods, the tumor segmentation ground truth is also shown in the last column for better view of the tumor preservation by different methods. The pix2pix can not well preserve the tumor information from T2, while our proposed methods generate images closer to the target modalities. The quantitative results are shown in Table \ref{psnr}. The proposed MGAN has a significant improvement compared to pix2pix in both PSNR and SSIM. Besides, with the introduction of multi-modality tumor consistency loss, TC-MGAN and TC-MGAN bw can more effectively preserve the tumor information and hence show higher SSIM scores.
      
\begin{table}[t]
\caption{Comparison of tumor segmentation accuracy.}
\label{dice}
\centering
\def\temptablewidth{0.478\textwidth}
{\rule{\temptablewidth}{1pt}}  

\begin{tabular*}{\temptablewidth}{@{\extracolsep{\fill}}cccc}

&\textbf{Method}&\textbf{DICE}&\\ 
\hline
&3*pix2pix&0.7381$\pm $0.3611&\\
\hline
&MGAN&0.7440$\pm ${0.3588}&\\
\hline
&TC-MGAN&\textbf{0.7585}$\pm $\textbf{0.3519}&\\ 
\hline
&TC-MGAN bw&0.7579$\pm $0.3513&\\
\hline \hline
&Only T2&0.7404$\pm $0.3600&\\ 
\hline
&4 Modalities&0.8142$\pm $0.3003&\\

\end{tabular*}
{\rule{\temptablewidth}{1pt}}
\end{table}



\subsection{Tumor Segmentation Results}
\label{ssec:subhead_seg}
In Table \ref{dice}, we use the synthesized images by different methods to boost the tumor segmentation accuracy, we also compare the DICE scores with only using T2 (Only T2) and using whole four real modalities (4 Modalities). MGAN generates higher segmentation accuracy than baseline model pix2pix. Our proposed TC-MGAN shows the best segmentation improvements to Only T2 and is the closet to the result when four real modalities are available. TC-MGAN bw shows a slightly lower segmentation score due to its lower synthesized image quality on FLAIR modality.

\section{CONCLUSION}
\label{sec:typestyle}

In this paper, we propose TC-MGAN to synthesize three missing brain MR modalities from a single modality. Based on the pix2pix, we introduce modality labels and multi-modality tumor consistency loss to enhance the synthesis quality and preserve tumor information from source modality. Our experiments show that the proposed method can be used not only to generate high-quality multi-modality MR images, but also as a data augmentation method to improve the performance of the segmentation network. In the future, more clinically meaningful evaluation will be conducted and we will further explore the application of this method in multi-modality medical image synthesis.


\vfill
\pagebreak



\clearpage
\bibliographystyle{IEEEbib}
\bibliography{main}

\end{document}